\documentclass[sigconf, nonacm]{acmart}

\newcommand\vldbdoi{XX.XX/XXX.XX}
\newcommand\vldbpages{XXX-XXX}
\newcommand\vldbvolume{19}
\newcommand\vldbissue{1}
\newcommand\vldbyear{2025}
\newcommand\vldbauthors{\authors}
\newcommand\vldbtitle{\shorttitle} 
\newcommand\vldbavailabilityurl{}
\newcommand\vldbpagestyle{plain}

\usepackage{subcaption}
\usepackage[utf8]{inputenc}
\usepackage{pdflscape}  
\usepackage{longtable}  
\usepackage{booktabs}   
\usepackage{array}      



\newif\ifdraft
\draftfalse                                                                              
\ifdraft
  \newcommand{\note}[1]{{\textcolor{red}{\emph{#1}}}}
  \newcommand{\ian}[1]{{\textcolor{blue}  { ***Ian:      #1 }}}
  \newcommand{\arham}[1]{{\textcolor{purple}  { ***Arham:      #1 }}}
    \newcommand{\kyle}[1]{{\textcolor{green}  { ***Kyle:      #1 }}}

\else
  \newcommand{\note}[1]{}
  \newcommand{\ian}[1]{}
  \newcommand{\arham}[1]{}
    \newcommand{\kyle}[1]{}

\fi

\newcolumntype{L}[1]{>{\raggedright\let\newline\\\arraybackslash\hspace{0pt}}p{#1}}

\begin{document}
\title{LSHBloom: Internet-Scale Text Deduplication}

\author{Arham Khan}
\orcid{0009-0006-6960-6651}
\affiliation{%
  \institution{University of Chicago}
  \city{Chicago}
  \state{Illinois}
}
\email{arham@uchicago.edu}

\author{Robert Underwood}
\affiliation{%
  \institution{Argonne National Laboratory}
  \city{Lemont}
  \country{Illinois}
}
\email{runderwood@anl.gov}

\author{Carlo Siebenschuh}
\affiliation{%
  \institution{University of Chicago}
  \city{Chicago}
  \state{Illinois}
}
\email{siebenschuh@cs.uchicago.edu}

\author{Yadu Babuji}
\affiliation{%
  \institution{University of Chicago}
  \city{Chicago}
  \state{Illinois}
}
\email{yadunand@uchicago.edu}

\author{Aswathy Ajith}
\affiliation{%
  \institution{University of Chicago}
  \city{Chicago}
  \state{Illinois}
}
\email{aswathy@uchicago.edu}

\author{Kyle Hippe}
\affiliation{%
  \institution{University of Chicago}
  \city{Chicago}
  \state{Illinois}
}
\email{hippekp@uchicago.edu}

\author{Ozan Gokdemir}
\affiliation{%
  \institution{University of Chicago}
  \city{Chicago}
  \state{Illinois}
}
\email{ogokdemir@uchicago.edu}

\author{Alexander Brace}
\affiliation{%
  \institution{University of Chicago}
  \city{Chicago}
  \state{Illinois}
}
\email{abrace@cs.uchicago.edu}

\author{Kyle Chard}
\affiliation{%
  \institution{University of Chicago}
  \city{Chicago}
  \state{Illinois}
}
\email{chard@uchicago.edu}

\author{Ian Foster}
\affiliation{%
  \institution{Argonne National Laboratory}
  \city{Lemont}
  \country{Illinois}
}
\email{foster@anl.gov}

\begin{abstract}
Modern internet-scale text databases require robust entity deduplication to maintain data integrity and reduce processing overhead. While Locality-Sensitive Hashing (LSH) is the established paradigm for near-duplicate detection, traditional LSHIndex structures based on trees or hashmaps suffer from large disk footprints and pointer-chasing latencies. We propose LSHBloom, a novel architectural extension to MinHashLSH that replaces traditional indexing structures with an array of independent probabilistic Bloom filters. By mapping LSH bands directly to contiguous bit vectors, optimizing for spatial locality, and co-designing our software with modern supercomputer hardware, LSHBloom achieves $12\times$ higher throughput than standard MinHashLSH on the 39M-document peS2o dataset. Our design reduces disk footprint by $18\times$ while maintaining near-identical recall and precision. We provide a formal analysis of the analytically boundable error rates and demonstrate, through extrapolation, that LSHBloom enables billion-scale record deduplication on commodity hardware.
\end{abstract}

\maketitle

\pagestyle{\vldbpagestyle}
\begingroup\small\noindent\raggedright\textbf{PVLDB Reference Format:}\\
\vldbauthors. \vldbtitle. PVLDB, \vldbvolume(\vldbissue): \vldbpages, \vldbyear.\\
\href{https://doi.org/\vldbdoi}{doi:\vldbdoi}
\endgroup
\begingroup
\renewcommand\thefootnote{}\footnote{\noindent
This work is licensed under the Creative Commons BY-NC-ND 4.0 International License. Visit \url{https://creativecommons.org/licenses/by-nc-nd/4.0/} to view a copy of this license. For any use beyond those covered by this license, obtain permission by emailing \href{mailto:info@vldb.org}{info@vldb.org}. Copyright is held by the owner/author(s). Publication rights licensed to the VLDB Endowment. \\
\raggedright Proceedings of the VLDB Endowment, Vol. \vldbvolume, No. \vldbissue\ %
ISSN 2150-8097. \\
\href{https://doi.org/\vldbdoi}{doi:\vldbdoi} \\
}\addtocounter{footnote}{-1}\endgroup

\ifdefempty{\vldbavailabilityurl}{}{
\vspace{.3cm}
\begingroup\small\noindent\raggedright\textbf{PVLDB Artifact Availability:}\\
The source code, data, and/or other artifacts have been made available at \url{\vldbavailabilityurl}.
\endgroup
}

\section{Introduction}

Modern text databases must handle numerous texts at the internet-scale, particularly with the advent of data-hungry systems like large language models (LLMs) that require access to many millions of documents. Redundant text records are a major source of inefficiency in these systems---recent studies have demonstrated that large web-scraped corpora such as CommonCrawl \cite{nagel2023common} (roughly 300B webpages) are comprised of between 14\% to 52\% near-duplicate records \cite{frobe2021copycat}. Many modern applications that rely on text databases benefit strongly from procedures that mitigate data redundancy including 1) internet-scale LLM pre-training where data redundancy has marked negative effects on language modeling ability~\cite{tan202415pintstechnicalreportpretraining, lee2021deduplicating} and significantly inflates training costs~\cite{chen-etal-2025-revisiting}, 2) retrieval-augmented generation systems that utilize vector databases where storing $O(d)$ fewer documents results in $O(log(D))$ faster query times \cite{malkov2018efficient} and mitigates the substantial GPU cost involved in embedding text for insertion \cite{gao2023retrieval}, 3) full-text search systems where storing $O(d)$ fewer documents results in a linear decrease in storage costs and in query complexity \cite{zobel2006inverted, aravind2024innovative}, 4) database tasks such as topic modeling or clustering where $O(d)$ fewer documents translates into an $O(d)$ speedup \cite{blei2003latent, grootendorst2022bertopic}, and 5) lookup tasks, such as those in search engines, that benefit from operations that can group text data by near-duplicates to group or distill query results (e.g., near-duplicate detection, fuzzy grouping) \cite{charikar2002, andoni2018approximate, broder1997resemblance}. 

Traditional relational entity resolution techniques based on exact-matching or common field values (e.g., blocking) \cite{christen2012data} fall short for text due to parsing noise (aberrations in raw text content introduced when parsing from PDF, XML, HTML, etc. into raw text) and the frequent absence or unreliability of metadata when obtaining documents from the web. Consequently, recent work has focused on the use of approximate similarity metrics such as n-gram overlap~\cite{soldaini_dolma_2024, li2024datacomp} and the Jaccard similarity~\cite{broder1997resemblance, broder1998min}.

The computational infeasibility of performing pairwise comparisons between documents in large text datasets motivated the development of MinHashLSH~\cite{leskovec2014finding}, which applies a locality-sensitive hashing (LSH) scheme to avoid comparing dissimilar documents. MinHashLSH is by far the most popular method for text deduplication in LLM data ingestion workflows \cite{brown2020language, rae2021scaling, olmo2025olmo, touvron_llama_2023, dubey_llama_2024, together2023redpajama, gao2020pile800gbdatasetdiverse, mireshghallah2023smaller, black-etal-2022-gpt, biderman2023pythia, penedo2025fineweb2pipelinescale, penedo2023refinedweb, li2023starcoder, lozhkov2024starcoder}; however, the associated MinHashLSH index can become slow and grow extremely large when applied at internet scale, presenting bottlenecks in throughput as well as in memory and disk resources. For example, applying MinHashLSH to the tiny peS2o dataset \cite{peS2o} containing 39M academic documents takes more than 35 core hours on the Polaris supercomputer and requires more than 200~GB of disk space (see \autoref{sec:scaling}). Overcoming these bottlenecks is paramount: language models are routinely retrained with newly acquired data \cite{dai2024llms, parmar2024reuse} and vector databases are continually updated with new content \cite{taipalus2024vector, ma2023comprehensive}. Continuously updating these collections requires efficient data ingestion, cleaning, and deduplication workflows. Moreover, when constructing text datasets, upstream changes in data pipelines such as the mode of data ingestion, cleaning, or preprocessing may alter the representation of existing documents and therefore, necessitate rerunning deduplication workflows on large corpora to mitigate data redundancy. Beyond a certain size, the runtime and disk requirements imposed by existing state-of-the-art deduplication techniques, such as MinHashLSH, render these processes infeasible to run.

To address these shortcomings, we introduce a new method, LSHBloom, which significantly improves upon the state-of-the-art in terms of the computational costs required to achieve a desired level of deduplication performance.
The key insight here is that the process that MinHashLSH employs to find matches between documents can be approximated, with a modest and analytically \textit{bound-able} loss of accuracy by using Bloom filters. This use of Bloom filters enables LSHBloom to achieve $12\times$ higher throughput than MinHashLSH by interfacing with inexpensive bit arrays rather than inserting and querying against a comparatively slow tree or hashmap-based LSHIndex. Most importantly, this use of Bloom filters means that LSHBloom requires 18$\times$ less disk space than MinHashLSH to deduplicate peS2o~\cite{peS2o}. We demonstrate that this space advantage remains as the number of documents increases and makes it feasible for LSHBloom to process several billion documents with significant space savings without sacrificing deduplication quality. Some researchers have experimented with using both MinHashLSH and Bloom filters to perform deduplication. For instance, the RefinedWeb~\cite{penedo2023refinedweb}, RedPajama~\cite{together2023redpajama}, and Gopher~\cite{rae2021scaling} datasets utilize Bloom filters for exact matching before finding near-duplicates using MinHashLSH (i.e., they use these techniques in \textit{conjunction}). By contrast, LSHBloom \textit{replaces} the traditional tree or hashmap-based index for near-duplicate detection in MinHashLSH with a space-efficient index comprised of Bloom filters that is co-designed with modern supercomputing hardware to enhance throughput.

LSHBloom is designed as a drop-in replacement for the industry-standard MinHashLSH~\cite{brown2020language, olmo2025olmo, dubey_llama_2024, rae2021scaling}---preserving deduplication fidelity while providing significant headroom in scaling capacity as the cardinality of text datasets increases. Historical analysis of frontier models indicates that LLM training dataset sizes almost quadruple each year \cite{epoch2024datasetsizetrend}. Beyond the increasing availability of large-scale general-purpose web corpora \cite{gao2020pile800gbdatasetdiverse, penedo2023refinedweb, penedo2025fineweb2pipelinescale} (including CommonCrawl \cite{nagel2023common}, which adds 3-5B newly scraped web pages each month), coordinated national initiatives such as the Genesis Mission \cite{EO14363} will integrate decades of scientific research into new, large-scale datasets for LLM training. This trend towards larger text datasets is further exacerbated by the recent strategic shift towards using high-volume synthetic data generation to continue scaling LLM training \cite{gunasekar_textbooks_2023, javaheripi_surprising_2023} despite limitations in the amount of available human-authored data. The ease with which one can generate vast quantities of synthetic text contributes to an ever-increasing data management burden. These observations motivate the development of LSHBloom, a robust data deduplication technique for large text datasets that is effective, efficient, and scalable well into the future. In summary, our contributions are:

\begin{enumerate}
\item We present the first side-by-side benchmark comparison of state-of-the-art text deduplication algorithms utilized by LLM data processing pipelines, comparing both their deduplication performance (in terms of precision, recall, and F1) as well as their resource requirements as we increase the size of the underlying text data.
\item We present an early study on the effects of tuning the hyperparameters for each of these algorithms, and identify the best parameter settings for each.
\item We carefully co-design the implementation with the modern supercomputer hardware to maximize throughput.
\item We develop LSHBloom, a novel text deduplication algorithm that outperforms or matches the existing state-of-the-art in deduplication performance while requiring significantly fewer resources and therefore enhancing scalability for the internet-scale data processing regime.
\end{enumerate}

\section{Background}

We first introduce the deduplication problem and describe the state-of-the-art for large-scale text deduplication pipelines.

\subsection{Defining the Deduplication Problem for Text}

In the context of database systems, text deduplication is a specialized form of Online Duplicate Elimination. Unlike batch-oriented entity resolution, which materializes all pairwise similarities within a static table \cite{christen2012data}, text deduplication in modern ingestion pipelines functions as a Streaming Approximate Membership Query (SAMQ). The objective is to determine, in real-time, if an incoming record represents a redundant entry relative to the existing corpus state.

 Formally, we define this task as a set-similarity membership problem. Suppose a similarity function $S(d_i, d_j): D \times D \rightarrow [0,1]$ that takes as input two documents and outputs a similarity score between 0 and 1. We define our task as follows: Given a stream of $N$ documents, $D_N =$ \{$d_1$, ..., $d_N$\}, and a similarity threshold, $T \in [0, 1]$, for each document, $d_i$, evaluate the binary decision function $\mathcal{F}(d_i)$ which returns 1 if $\exists d_j \in D_{seen} : S(d_i, d_j) \geq T$ where $D_{seen} = \{d_j \mid j < i\}$ and 0 otherwise. 
 This formulation prioritizes ingestion throughput and state compression as storing the full history, $D_{seen}$, is intractable at scale.




The choice of similarity function, $S$, is crucial for maximizing join fidelity. For example, one could simply stipulate that two documents must be byte-for-byte identical. 
However, this definition may exclude many document pairs that are not exact matches, but have high overlap in their content, as are encountered frequently in practical text ingestion pipelines. For example, the same document may be available from two separate venues (say, a pre-print on Arxiv and a camera-ready conference publication) that differ only slightly due to editorial decisions. Also, when web-scraping text or obtaining articles from publishers, text data may be obtained in varied formats (e.g., HTML, PDF, XML), and the appropriate format-specific text parsing pipelines will produce small differences in raw text content even when applied to the same underlying document. 
PDF parsers that rely on optical character recognition (OCR) can also introduce artifacts that render exact matching woefully ineffective. As a result, properly deduplicating text data requires adopting approximate similarity measures that are robust to minor aberrations.

\subsection{Approximate Similarity Metrics for Text}

One way to define an approximate similarity measure between documents is to measure the proportion of n-grams in that document that were previously seen in the data stream. Popular techniques such as those used to build the DOLMA~\cite{soldaini_dolma_2024} dataset and DataComp-LM~\cite{li2024datacomp} baseline dataset employ this procedure to identify duplicates. However, these techniques can be sensitive to the relative frequency of n-grams within a document or dataset.

Alternatively, some techniques, such as MinHashLSH, take a set-theoretic perspective on document similarity.
In this set-theoretic paradigm, a text document is viewed as a set of n-grams, and the similarity between two such sets, $A$ and $B$, is defined in terms of the Jaccard similarity: $J(A, B) = \frac{|A \cap B|}{|A \cup B|}$.




However, computing the Jaccard similarity requires evaluating set membership for each item present in the union of both sets. For extremely high-dimensional objects such as documents, computing the Jaccard similarity explicitly can be extremely expensive. The MinHashing procedure \cite{broder1998min} provides a means of efficiently approximating the Jaccard similarity of two documents. 
Rather than compare the elements of each set directly, this procedure randomly permutes the order of set elements to produce so-called MinHash signatures---lossy, fixed-length representations of the underlying high-dimensional objects. The probability that the MinHash signatures for two sets are equal is exactly equal to the Jaccard similarity of the two sets \cite{broder1998min, leskovec2014finding}. Therefore, we can repeatedly apply the MinHashing procedure to any two documents and compare the resulting hashes for equality to obtain an estimate of their Jaccard similarity. This estimate grows more accurate as we increase our sample size (the number of random permutations).

\subsection{MinHashLSH for Text Deduplication}
\label{sec:minhash_lsh_bg}

The use of MinHash-estimated Jaccard similarity, $J_H$, as an approximation for Jaccard similarity, $J$, provides for more efficient similarity computations because the procedure employed to estimate the similarity of two documents has linear complexity in the number of random permutations applied, rather than quadratic complexity in the number of n-grams present in each document.
However, the need to compare every pair of MinHash signatures for similarity during deduplication still introduces quadratic cost-scaling with respect to the number of documents in our dataset.

Fortunately, there exists a technique that allows us to compare only those pairs with a high likelihood of similarity, rather than all pairs. MinHashLSH~\cite{leskovec2014finding} applies a locality-sensitive hashing scheme for MinHash signatures to avoid comparing dissimilar documents. Given a set of MinHash signatures, one can construct a MinHash signature matrix in which each column contains all MinHash values for a single document. Then, given a desired Jaccard similarity threshold, $T$, one can determine an optimal band size, $b$, with which to group the rows of the MinHash signature matrix. This results in $b$ groupings of rows (bands) of size $r$. With high probability, if two bands of the MinHash signature matrix are exactly identical for two documents (columns), then those two documents have Jaccard similarity greater than or equal to $T$. This procedure enables us to produce a series of candidate pairs of documents that we can either take as duplicate pairs (our approach in this work) or further evaluate for similarity. This method avoids pairwise comparison between every pair of documents in the corpus but introduces 
errors due to the use of locality-sensitive hashing.

\begin{figure}[ht!]
    \centering
    \includegraphics{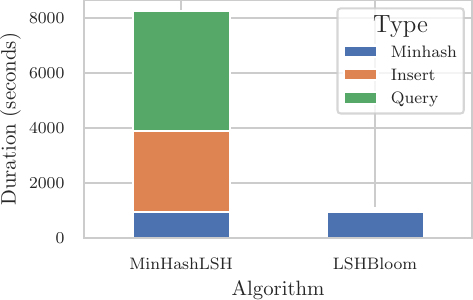}
    \caption{Breakdown of wall clock time on 10\% of peS2o for conventional MinHashLSH and our LSHBloom method}
    \label{fig:wall_clock_time_breakdown}
\end{figure}

One can implement a prefix-tree or a hashmap to search efficiently for candidate duplicate documents when evaluating a new document for duplication. However, this requires storing a representation of each document, and while MinHash representations are small compared to raw documents, the MinHashLSH index becomes large for large datasets: e.g., 277~TB for 5B documents. Moreover, navigating these data structures may require pointer-chasing and, therefore, non-contiguous memory accesses that cause cache misses. Inserting into and querying against these indices constitutes a significant portion (>85\%) of algorithm runtime, acting as the primary bottleneck for real-time stream throughput at extreme scales (see \autoref{fig:wall_clock_time_breakdown}).  Here, we remedy this space and throughput issue by employing a novel combination of MinHashLSH and Bloom filters.

\subsection{Bloom Filter}

The Bloom filter \cite{bloom1970space} is a space-efficient probabilistic data structure used to test whether an element is a member of a set. It can determine with certainty if an element is not in the set, but if it indicates the element is in the set, there is a small probability of error. That is, the Bloom filter may admit false positives but not false negatives.

A Bloom filter consists of a contiguous bit array of size $m$, with all bits initially set to 0. 
$k$ independent hash functions are defined, each of which hashes an element to an integer in the range 1 to $m$.
To insert an element into the Bloom filter, it is passed through the $k$ hash functions, and the resulting $k$ positions in the bit array are set to 1.
We can then check if an element is in the set by hashing using the same $k$ hash functions.
If all $k$ of the corresponding positions in the bit array are 1, the filter returns a positive. If any of those $k$ bits is 0, the element is not contained in the set.

Bloom filters have been used extensively in distributed computing for a variety of purposes \cite{broder2004network,chervenak2008globus,tarkoma2011theory}.

\section{Related Work}

\ian{To cite perhaps: \cite{tirumala2023d4,he2024softdedup,muennighoff2023scaling,chang2024scaling}.}

The task of streaming text deduplication lies at the intersection of the literature on Set-Similarity joins, Approximate Membership Queries, and Streaming Entity Resolution. While traditional approaches prioritize fidelity at the cost of computational resources, LSHBloom targets the Pareto-optimal frontier of high-dimensional streaming deduplication---offering state-of-the-art deduplication fidelity while minimizing resource usage.

\subsection{Similarity Joins}

Traditional techniques for performing a Set-Similarity Join attempt to identify all pairs of records such that their set-similarity (as measured by the Jaccard, Dice, or Cosine metrics) exceeds a given threshold. \citet{mann2016empirical} perform a comprehensive evaluation of techniques for performing the Set-Similarity Join (e.g., AllPairs, PPJoin). While employing certain filter optimizations can improve upon the quadratic cost of computing exact set-similarity metrics, ultimately, these techniques are not appropriate for extremely high-dimensional entities such as documents. Techniques such as SemDedup~\cite{abbas2023semdedup} aim to deduplicate text entries based on their overlap in terms of semantic content rather than using purely textual overlap to compute similarity; however, due to the use of language models, this technique incurs significant computational costs. Metadata-based approaches (e.g., exact URL matching, matching DOIs), including blocking using common field values \cite{christen2012data},  are a simple and efficient alternative, but are applicable only when a reliable source of metadata exists.  Unfortunately, for our target datasets
metadata is often unavailable or is untrustworthy, as when, for example:
1) a document appears in two or more collections with different metadata schemes (e.g., ArXiV vs.\ ACM, different GitHub repositories);
2) sections/paragraphs of text are copied, as in extended journal versions of computer science papers; or
3) metadata is not curated (e.g., an ArXiV entry may not be updated to reflect a subsequent publication).

\subsection{Approximate Membership Queries}

Rather than attempting to directly identify pairs of similar entities, some techniques instead attempt to simply determine, online, whether a new record already belongs to the existing dataset at the time of ingestion. For example, by using a special data structure such as a hashmap or Bloom filter~\cite{bloom1970space} to query for matches against previously ingested records. While naive exact-matching with hashmaps or Bloom filters is efficient, it is not robust to the minor variations in text content that are inherent when using text data obtained from the web or parsed from various formats. 

For text, it is preferable to use approximate notions of similarity. For instance, the Streaming Quotient Filter (SQF)~\cite{dutta2013streaming} is an approximate duplicate detection approach for streams of incoming data. However, the SQF is optimized for approximate duplicate detection in low-dimensional spaces and is not appropriate for text deduplication. For text, techniques such as MinHashing~\cite{broder1998min} and MinHashLSH~\cite{leskovec2014finding} are more appropriate due to their ability to compress high-dimensional text records into fixed-length representations. The use of LSH functions in MinHashLSH has the additional advantage of identifying candidate duplicates using hashing as opposed to running comparisons with each existing record in the dataset. MinHash-based techniques use the Jaccard similarity to measure text overlap; however, n-gram overlap is also a reasonable metric---albeit one that is sensitive to the frequency of n-grams in a given document. Frequently, techniques evaluate the n-gram overlap of a document with the n-gram content of the remainder of the corpus using a Bloom filter, trading some granularity of comparison for space efficiency and increased throughput. This approach was utilized in preparing the DOLMA~\cite{soldaini_dolma_2024} dataset and the DataComp-LM~\cite{li2024datacomp} baseline dataset. LSHBloom falls in this category of techniques for online approximate membership queries.

\subsection{State-of-The-Art Deduplication Techniques for Text}\label{sec:techniques}

Below, we describe in detail popular techniques for identifying duplicates in text datasets, commonly applied in real-world text dataset preparation. 

\textbf{CCNet} \cite{wenzek2020ccnet} is an automated preprocessing pipeline that aims to extract high-quality monolingual language datasets from the Common Crawl 
corpus of Web content. CCNet was employed as part of the data processing workflow for producing the Llama series of models \cite{touvron_llama_2023, touvron_llama_2023-1, dubey_llama_2024}. CCNet performs dataset sharding, deduplication, language identification, and quality filtering. CCNet's deduplication process begins by preprocessing text so that all characters are lowercase and eliminating special Unicode characters. CCNet then splits documents by newline characters, computes SHA1 hashes for each resulting unit of text, and compares SHA1 hashes between units of text to identify duplicates. CCNet is only capable of identifying documents that are exactly identical to one another. 

\textbf{Dolma} \cite{soldaini_dolma_2024} deduplicates paragraphs by exact matching using a Bloom filter. 
The Dolma-Ngram alternative splits paragraphs into n-grams and checks whether n-grams are duplicated by querying against a Bloom filter. If the proportion of duplicated n-grams in a paragraph exceeds a threshold, $T \in$~[0,~1], then that paragraph is marked as a duplicate. By default, $T$~=~1.0 in their configuration. These two techniques were used to produce the Dolma dataset~\cite{soldaini_dolma_2024} used to produce the Olmo series of language models \cite{groeneveld-etal-2024-olmo, olmo20242}.

\textbf{DataComp-LM} \cite{li2024datacomp} (DCLM) deduplicates items at the document and paragraph levels simultaneously by using a Big Friendly Filter, an extension of the Bloom filter. This technique was applied to create the DataComp-LM baseline dataset. DCLM splits each document into paragraphs on newline characters and tokenizes paragraphs with the UniSeg tokenizer \cite{ye2023uniseg}. For each paragraph, it queries each n-gram against its Bloom filter and removes duplicated n-grams. If the proportion of duplicated n-grams in a paragraph exceeds a user-defined threshold, $T \in$ [0,~1], the entire paragraph is removed from the document. If the proportion of duplicated n-grams in the entire document exceeds the user-defined threshold, then the document is removed from the dataset. They employ the same threshold parameters for paragraph-level and document-level deduplication. 

\textbf{MinHashLSH} \cite{leskovec2014finding} the MinHashLSH method as described in \autoref{sec:minhash_lsh_bg}. MinHash-based deduplication was applied in the data ingestion pipeline for OpenAI's GPT-3~\cite{brown2020language}, DeepMind's Gopher~\cite{rae2021scaling}, AllenAI's Olmo3~\cite{olmo2025olmo}, and Meta's Llama3~\cite{dubey_llama_2024}. MinHash-based deduplication and MinHashLSH were also employed in creating many popular language modeling datasets, such as RedPajama~\cite{together2023redpajama}, the Pile~\cite{gao2020pile800gbdatasetdiverse} (used for GPT-J~\cite{mireshghallah2023smaller}, GPT-NeoX~\cite{black-etal-2022-gpt}, and Pythia~\cite{biderman2023pythia}), FineWeb2~\cite{penedo2025fineweb2pipelinescale}, RefinedWeb~\cite{penedo2023refinedweb} (used for the Falcon series of models), StarCoder~\cite{li2023starcoder}, and StarCoder2~\cite{lozhkov2024starcoder}. The sheer number of LLM data pipelines that employ MinHashLSH is a testament to its effectiveness. Furthermore, it highlights the potential impact of enhancing its scalability for internet-scale datasets through innovative techniques, such as LSHBloom.

\section{Our Method: LSHBloom}


We propose a memory-efficient alternative to MinHashLSH for document deduplication. 
Rather than employ the traditional prefix-tree or hashmap index used to find exact matches between the MinHash signatures in each band, we develop a novel Bloom filter index structure for MinHashLSH. We instantiate a series of Bloom filters, one corresponding to each band of the MinHash signature matrix. By inserting several rows of MinHash signatures into a fixed number of Bloom filters of fixed size, we achieve a large reduction in space usage by the index and a significant speedup in duplicate searches. The size of each Bloom filter and the number of hash functions utilized are chosen according to the method determined by \citet{bender2022optimal}, which guarantees optimal (minimal) space usage. LSHBloom effectively acts as a Similarity-Aware Approximate Membership Query data structure, combining the compressive properties of MinHashLSH with the probabilistic indexing of Bloom filters to create a high throughput, space-efficient, online deduplication method for text ingestion workflows.

\subsection{Insertion}
\label{sec:method_insert}
Given a Jaccard similarity threshold $T \in$ [0, 1], MinHashLSH traditionally groups signatures into a series of $b$ bands of size $r$ such that an exact match in one band for two columns of the MinHash signature matrix denotes a duplicate document. That is, with high probability, the two documents with a matching band have Jaccard similarity greater than or equal to the threshold. To avoid storing the signatures in full, as is required by MinHashLSH, we assign each band of the minhash signature matrix %
its own Bloom filter. Thus, we have in total $b$ Bloom filters, and a collision in any one indicates a duplicate document. 

To insert a document into the index, we first compute its MinHash signature, which is a vector of integers (each produced by a universal hash function). Then, we group the rows into $b$ bands of size $r$, as dictated by the MinHashLSH algorithm. 
It is desirable to hash this list of integers into a single value that we can readily insert into a Bloom filter in a manner that minimizes collisions. 
To do so, we take advantage of the fact that each MinHash signature value ($h_i(x_i)$) is computed using a universal hash function (in our case, SHA1~\cite{sha1}), with the property that the probability of a hash collision is $1/N$ where $N$ is the range of possible hash values. Then, we can hash a vector $\bar{x}$ of $r$ integers as follows \cite{carter1977universal}:
\[
    h(\bar{x}) = \left(\sum_{i=1}^{r} h_i(x_i)\right) \mathrm{mod}\  {N}
\]


This approach enables us to map each band of MinHash signatures to a single integer value with known collision probability ($1/N$). To insert a document into the index, we simply insert each of the $b$ final integer values into each of the $b$ Bloom filters corresponding to the $b$ bands in the signature matrix. 

\subsection{Querying}
Querying a document against the index is a similar process to insertion. 
We compute the document's MinHash signature, group the rows into bands, and then hash each band as described in \autoref{sec:method_insert}.
If there is a collision for any band of signatures, we mark that document as a duplicate. 
This approach mimics the original MinHashLSH procedure and gives the same probabilistic guarantees with respect to the Jaccard similarity of candidate pairs, with a slight increase in false positive probability due to our use of Bloom filters. However, the false positive probability is configurable and can be set to be arbitrarily low. An additional benefit of LSHBloom is that the false positive overhead is analytically boundable, as we demonstrate below.

\subsection{Error Rate}

In traditional MinHashLSH, a collision in any band causes a document to be marked as a candidate duplicate. This is also the case in LSHBloom, where the use of Bloom filters incurs additional false positive overhead. Here, we analyze the error rate of LSHBloom. We define a false positive as a non-duplicate document that is incorrectly identified as a duplicate by LSHBloom; a false negative is a duplicate document that is not marked as a duplicate by LSHBloom. 

Given a Minhash signature matrix with $b$ bands with $r$ rows each corresponding to $b$ Bloom filters with their individual false positive rates set to $p$, the net false positive overhead of using Bloom filters will be $p\textsubscript{effective} = 1-(1-p)^b$. This is because a single collision in any band constitutes a duplicate, and false positive rates are independent for each Bloom filter. Therefore, the probability that no Bloom filter reports a false positive is $(1-p)^b$, and the overall false positive rate overhead introduced by using Bloom filters is the complement of this value. This means we can enforce a false positive overhead of $p\textsubscript{effective}$ by setting the value of $p$ for each Bloom filter as $p=1-(1-p\textsubscript{effective})^{1/b}$

The relationship between the false positive rate of the Bloom filter and the overall error rate of LSHBloom can be understood as follows. \citet{leskovec2014finding} determine that the false positive and false negative rate of MinHashLSH can be given by the following integrals, where $T$ is the Jaccard similarity threshold, $b$ is the number of bands in the MinHash signature matrix, and $r$ is the number of rows in each band:
\begin{equation}
    FP_{lsh} = \int_{0}^{T} 1 - (1 - t^r)^b dt
\end{equation}
\begin{equation}
    FN_{lsh} = \int_{T}^{1} 1 - (1 - (1 - t^r)^b) dt
\end{equation}

\citet{zhu_lshensemble} describe how to tune $b$ and $r$ to minimize these errors in MinhashLSH. We follow this approach in our work.

A Bloom filter will not yield false negatives, but may yield false positives. If MinHashLSH yields a false positive (by “mistakenly” yielding two identical bands), we get a false positive result. If MinHashLSH does not yield a false positive, the Bloom filter index may still yield a false positive result with probability $p$\textsubscript{effective}. Additionally, there is a small probability of hash collision ($1/N$) when reducing each band to a single integer (as described in \ref{sec:method_insert}) for insertion into the Bloom filter. Therefore, the overall false positive rate for LSHBloom is given by:
\begin{equation}
    FP_{\textrm{bloom}} = FP_{lsh} + (1-FP_{lsh})*(p_{\textrm{effective}} + b/N)
\end{equation}

Note that $p\textsubscript{effective}$ can be set arbitrarily small in practice so that this overhead is negligible. For 32-bit hashvalues (the default in packages such as \verb|datasketch|) $N = 2^{32}$, so the collision probability for bands before insertion into the Bloom filter is also small. 

Meanwhile, for each false negative that MinHashLSH produces, $p\textsubscript{effective}$ percent of those false negatives will be marked as positives due to false positive hits in the Bloom filter index, and $b/N$ percent of them will be marked positive due to collisions when reducing the bands of the MinHash signature matrix for insertion. That is,
\begin{equation}
    FN_{\textrm{bloom}} = (1-(p_{\textrm{effective}} + b/N))*FN_{lsh}
\end{equation}

\subsection{Optimizations}
\label{sec:lshbloom_optimizations}

We codesign our implementation of LSHBloom with supercomputer hardware to maximize gains in throughput. We describe these optimizations in detail below.

\subsubsection{Hashing}

After profiling the original implementation of LSHBloom, we optimized the hashing operations used in LSHBloom with careful hardware and software co-design.
Our profiling revealed that hashing of the integer vectors described in \autoref{sec:method_insert} accounted for over 90\% of the time spent in the insert and query routines for LSHBloom.
The key reason for this was the use of an inefficient software extended-precision integer representation included in Python, which stores extended integers as base-10 strings \cite{python_numeric_types}. By contrast, NumPy offers fixed-precision, C-style integer representations as well as support for vectorized additions, which enhance throughput \cite{harris2020array, numpy_data_types}. However, adding hundreds of fixed-precision integers may lead to overflows, inflating the collision probability of hashvalues before insertion into our Bloom filters.
The hash of each MinHash permutation is 64 bits, so adding them requires at most 71 bits of unsigned precision to perform without overflow.
With this in mind, 128-bit arithmetic can be used to avoid integer overflow.
Modern Intel and AMD \texttt{x86\_64} CPUs implement 128-bit integer arithmetic using the \texttt{adc} \cite{guide2011intel} (''add with carry``) instruction with equivalents such as \texttt{addc.u64} on Nvidia PTX \cite{nickolls2008scalable, nvidia_ptx_isa} on other platforms.
Additionally, the modular division operation performed as the last step can be implemented either using a simple bit mask or by simply using the lower order half of the 128-bit register resulting from the add with carry operation.
Using Rust's standard extended precision support \cite{matsakis2014rust, rust_reference_numeric} makes expressing this operation easy and results in only 30 lines of optimized \texttt{x86\_64} assembly depending on architecture and loop unrolling decisions. We replace our Python logic for the integer vector hashing operation described in \autoref{sec:method_insert} with this Rust function.
This single optimized function is over $94\%$ faster than the existing Python implementation, resulting in a $11\times$ end-to-end wall clock improvement to LSHBloom.

\subsubsection{Parallelization and Shared Memory}

Parallelization is an obvious strategy for acceleration. As shown in \autoref{fig:wall_clock_time_breakdown}, MinHashing accounts for the majority of the runtime in LSHBloom. Thankfully, each document can be MinHashed independently and this step can easily be parallelized across multiple nodes using frameworks such as Parsl \cite{babuji2019parsl}. On single-node runs (as in our evaluation), we can employ multiprocessing to speed up this step. When finally writing to the Bloom filter index, we do so sequentially (without parallelism). This is because, at the time of insertion, we must also ensure that the current document is not a duplicate of a previously inserted one. Naively parallelizing this step would lead to errors in duplicate identification. Parallelizing regardless and insisting upon synchronizing Bloom filters across threads or nodes would introduce significant communication overhead. Despite this lack of parallelism, our optimized hashing routines still enable extremely fast insert and query operations. Furthermore, we take advantage of the tiny size of our Bloom filter indices by hosting our Bloom filters in node-local shared memory segments (via \texttt{/dev/shm}), allowing us to locate our index in DRAM with swap partitions on local SSDs. This further accelerates insert and query operations on the Bloom filter index.

\subsection{Advantages}
\label{sec:lshbloom_advantage}


LSHBloom allows for the space-efficient deduplication of extremely large corpora. This is due to the Bloom filter's ability to determine the set membership properties of objects (e.g., documents) of arbitrary size with a fixed number of bits. The size of the LSHBloom index only depends on the number of planned documents, $n$, the number of bands, $b$, in the MinHash signature matrix, and the desired false positive rate bound, $p$, for each Bloom filter. If we expect $n$ documents and desire a false positive bound of $p$, then each of our $b$ Bloom filters requires exactly
    $m = -n * log(p)/(log(2))^2$ \textrm{ bits}~\cite{bender2022optimal}.

This means that the size of our index is invariant to the size of MinHash hashvalues---we can increase the size of MinHash hashvalues (to say, a 64 or 128-bit integer) to enhance the accuracy of our Jaccard similarity estimate at no additional cost. Meanwhile, because the traditional MinHashLSH index must store these hashvalues for comparison, increasing the size of the MinHash hashvalues results in a linear increase in the size of the traditional LSHIndex. Enhancing the accuracy of our probabilistic estimate of the Jaccard similarity by increasing the number of MinHash permutations we apply will increase the size of the MinHash signature matrix, which will only change the size of the Bloom filter index if the number of bands, $b$, in the signature matrix increases. Luckily, the number of bands in the signature matrix is generally far fewer than the number of total permutations.

As an example, say we use a Jaccard similarity threshold $T$ of 0.8, and 128 random permutations for MinHashing. For these parameter settings, MinHashLSH creates nine bands (using the method described in \cite{zhu_lshensemble} to tune $b$ and $r$) for the MinHash signature matrix, meaning that we require nine Bloom filters for our index. For these settings, setting a conservative $p$\textsubscript{effective} = 1e-10 for a dataset of 10B documents would only require 590~GB of space. Meanwhile, traditional MinHashLSH would require 46~TB (nearly $80\times$ more space) to store the equivalent LSHIndex. 



While the use of Bloom filters to satisfy memory and throughput constraints introduces additional false positive errors, the rate of false positives and false negatives is analytically boundable. Moreover, the additional false positive error overhead scales with the number of documents we intend to insert and scales inversely with the size of each filter. Therefore, there is a trade-off between space efficiency and correctness that is mediated by the size of each filter. However, as demonstrated by our example above, this trade-off tends to be highly favorable even on extremely large datasets. Additionally, LSHBloom only introduces a single additional hyperparameter, the false positive rate for our Bloom filters, meaning that our hyperparameter tuning burden is comparable to that of MinHashLSH.


Beyond storage savings, the Bloom filter index yields substantial enhancements in ingestion throughput due to two primary factors. 1) Computational efficiency: insertion and querying against the index requires only a few independent hash operations and modular arithmetic, which is fast on modern hardware (see \autoref{sec:lshbloom_optimizations}), and 2) Cache-Aware Data layout: in contrast to the pointer chasing inherent in tree or chained-hashmap-based indices, LSHBloom's use of contiguously allocated bitarrays maximizes spatial locality and minimizes cache misses.

\arham{I threw a bunch of calculations/data points in the above final sentences, we should discuss which we want to present here - which would be best to talk about at this point in the paper?}
\kyle{why 1GB for 100M and then 1.05GB for 39M?} \arham{the difference is the false positive rates, that accounts for the difference - but to limit confusion I'll leave it for 100M for now}

\section{Evaluation}


In evaluating a deduplication method for a particular application, we are concerned with its deduplication fidelity, throughput, and disk space requirements. We measure deduplication fidelity via the precision and recall metrics.
Low \textit{precision}, i.e., excessive false positives, is harmful as it unnecessarily reduces dataset size and, in deep learning, limits an LLM's ability to learn the underlying distribution of the text. Low \textit{recall}, i.e., excessive false negatives, admits a great degree of duplicate content, with negative effects on LLM quality in pre-training applications. High throughput and a small disk footprint is critical due to the extreme size of modern LLM datasets. For example, peS2o~\cite{peS2o} contains roughly 39 million documents, DOLMA~\cite{soldaini_dolma_2024} 5B documents, SmolLM3's pretraining dataset has over 15B documents~\cite{smollm3_datasets_2025}, and RedPajama~\cite{together2023redpajama} has over 100B documents---scales at which naive pairwise comparisons are infeasible. Therefore, deduplicating at the billion-record scale with high fidelity is paramount. 

To assess deduplication fidelity, we 
tune LSHBloom and other state-of-the-art deduplication methods on a synthetic, balanced dataset of 24K PDFs containing labeled duplicates, we refer to this as our \textit{tuning dataset}. We then use the best parameter settings for each algorithm and compare them on a larger synthetic dataset of 50K PDFs which we term the \textit{testing dataset}. 
Without sufficient fidelity, the system risks either under-identifying duplicates, leading to decreased downstream model quality and unnecessary processing overhead, or over-identifying duplicates, which unnecessarily reduces training dataset volumes.
To measure performance, we consider the storage utilization and runtime of each method on a larger collection of 39M PDFs, peS2o~\cite{peS2o}. 
Without a sufficiently high throughput and sufficiently low disk footprint, it may not be possible to process large datasets on available hardware resources.


In the following subsections, we describe the details of our evaluation, beginning with the evaluation methodology, then proceeding to our evaluation of fidelity, performance, and finally the tradeoffs between performance and fidelity.

\subsection{Experimental Methodology}

We describe our experimental methodology, including testbed hardware, state-of-the-art methods against which we compare, evaluation metrics, and datasets.

\subsubsection{Testbed Hardware} 
We ran our experiments on the Polaris supercomputer, a 560-node HPE Apollo 6500 Gen 10+-based system. Each node is equipped with a 32-core AMD Zen 3 2.8 GHz processor, with 512~GB of DDR4 RAM, and 3.2~TB of local SSD: representative of nodes found in large-scale high-performance computing facilities used for deduplication tasks.

\subsubsection{Baselines}
We compare LSHBloom with state-of-the-art deduplication methods widely used in large-scale LLM training: MinHashLSH \cite{broder1997resemblance}, Dolma and Dolma-Ngram \cite{soldaini_dolma_2024}, CCNet \cite{wenzek2020ccnet}, and DataComp-LM \cite{li2024datacomp} (see~\ref{sec:techniques} for details). 
We normalized all implementations to Python implementations to avoid performance differences for common procedures due to implementation language. This also enabled us to normalize the Bloom filter modules for those techniques that use them to a fast C/Python-based implementation provided by \verb|pybloomfiltermmap3| \cite{sinha2025pybloomfilter}. LSHBloom is evaluated using the optimized hashing routine described in \autoref{sec:lshbloom_optimizations}.
Below, we describe any changes or configurations to the methods. 

\textbf{MinhashLSH} \cite{broder1997resemblance}: We configure traditional MinhashLSH as described in \autoref{sec:minhash_lsh_bg}. We use the implementation offered by the popular \verb|datasketch| package \cite{zhu2025datasketch}, which implements a hashmap-based LSHIndex. We compute MinHashes using multiprocessing for both MinHashLSH and LSHBloom.
\textbf{DOLMA}: We apply both standard Dolma and Dolma-Ngram methods to document-level deduplication. For Dolma, we determine whether a document is duplicated by identifying duplicated paragraphs and comparing the percentage of document text duplicated to the overlap threshold, $T$. For Dolma-Ngram, we split each document into n-grams, determine how many n-grams are duplicated according to the Bloom filter, and compare the percentage of n-grams duplicated within that document to the overlap threshold, $T$. Our implementation aims to faithfully extend Dolma's two primary deduplication methods to document-level deduplication. 
\textbf{CCNet} originally performs deduplication at the paragraph level; we apply it at the document level 
by measuring the proportion of duplicated paragraphs in any given document against some tolerance threshold, $T \in$ [0,~1]. 
We augment CCNet with multiprocessing where appropriate (e.g., when computing hashes over each document) for a fair comparison during our scaling tests.
\textbf{DataComp-LM}: For our evaluation, we compare only with their document-level deduplication procedure. 

For Dolma, Dolma-Ngram, and DCLM, we must give an expected n-gram/paragraph count for the corpus; choosing this count accurately is crucial to making a fair comparison because it influences Bloom filter size and error rates. Because counting the number of n-grams/paragraphs in a corpus is very expensive compared to counting documents (as is required for MinHash-based techniques), we compute an estimate of the n-gram/paragraph count in a corpus by randomly sampling $N$ = 1000 documents from the corpus, computing the mean n-gram/paragraph count from our sample, and multiplying the mean count by the number of total documents in that corpus to obtain an estimated total n-gram/paragraph count. 

\subsubsection{Metrics}

\textbf{Fidelity:} We require methods that can
distinguish meaningfully between duplicate and non-duplicate documents. Here, we consider a "positive" label to mean that a document is a duplicate of another document in the corpus. 
We want these methods to yield few false positives (to retain the maximal number of informative records), few false negatives (to reduce the prevalence of duplication), and to provide a reasonable combination of accuracy and performance.
Particularly for large-scale deep learning, retaining the maximal number of training examples is highly desirable---therefore, we want few false positives. Similarly, allowing a large number of duplicates into the training corpus is unacceptable as this affects the model's training efficiency and decreases model quality. 
To account for these considerations, we evaluate each deduplication technique based on its ability to mitigate both false positives and false negatives using its precision (the proportion of duplicate labels that are true duplicates), its recall (the proportion of duplicates in the dataset that we correctly identify), and its F1 score (the harmonic mean of precision and recall):

\vspace{-2ex}

\[
    F1 = \frac{TP}{TP + \frac{1}{2}(FP + FN)}
\]

\vspace{1ex}
\noindent
\textbf{Performance:}
To assess the resource costs of each technique, we measure wall clock time (using \verb|/user/bin/time|) and disk space requirements (using Python's standard \verb|os.path.getsize|).

\subsubsection{Constructing the fidelity evaluation dataset}
\label{sec:25k}


To permit the rigorous evaluation of de-duplication methods, we created a pipeline for generating labeled datasets of duplicate documents. \citet{siebenschuh2025adaparse} produced a dataset of parsed PDFs for evaluating their adaptive PDF parsing technique, AdaParse. We utilize this baseline dataset to construct synthetic tuning and testing datasets for the deduplication task.  
The baseline dataset consists of scientific articles with two versions: one scraped from HTML and the other parsed from PDF. These span diverse disciplines (mathematics, physics, chemistry, biology, economics, engineering, and medicine) as identified through keywords or classification algorithms based on metadata. 
Text content from HTML is directly extracted. Each PDF is parsed 
by applying a text extraction or OCR tool---one of 
PyMupDF \cite{PyMuPDF}, Nougat \cite{blecher2023nougat}, or Tesseract \cite{smith2007overview}, each used with roughly the same frequency. We construct datasets containing duplicated documents from this collection of parsed texts, either by including multiple versions of the same document parsed with different tools and/or by including randomly truncated versions of that document. We balance the number of parser and truncation-based duplicates in each dataset we construct, which prevents evaluation bias towards techniques that are better suited to identifying just one of these types of duplication. These scenarios reflect common forms of duplication encountered in real-world web scraping pipelines where the same document may be ingested via different text extraction tooling due to differences in document origin, the same document being obtained in multiple formats (HTML, XML, PDF, etc.), and due to parsing errors, which can abruptly skip or truncate portions of the document's text. 

We build a balanced dataset of 24,000 documents for use in tuning hyperparameters for each deduplication algorithm, the \textit{tuning dataset}. We further evaluate each technique using its best hyperparameter settings on datasets of 50,000 documents with increasing levels of duplication, from 10\% to 90\% duplicate instances, the \textit{testing datasets}.


\subsubsection{Choice of values for Tuning Parameters}

We evaluate each technique for a range of parameter settings. Specifically, for each threshold parameter, such as the Jaccard similarity threshold for MinHashLSH and the text overlap threshold for DataComp-LM, we evaluate values from 0.2 to 1.0 in steps of 0.2. We vary the number of permutations for MinHashing from 32 to 256 by powers of two. For n-gram techniques, we explore n-gram sizes 1, 2, 5, 7, 13, and 26. For MinHashLSH and LSHBloom, we also gathered results for $T$ = 0.5 and 48 random permutations, as the surrounding parameter values demonstrated sharp jumps in F1 score, and we wanted to explore this change in the score at a finer granularity. All Bloom filters (for DOLMA/DOLMA-ngram, DataComp-LM) are assigned a false positive rate $p$ of 1e-5. For LSHBloom, we set $p\textsubscript{effective}$ to 1e-5.

 
 It is possible that one could properly tune these deduplication methods further using more incremental changes in their parameter settings.
 However, the expensive runtime of n-gram techniques on large corpora, coupled with the need to comb over parameter settings in small increments points to the inherent difficulty and cost associated with properly tuning these methods for maximal deduplication performance.
 Meanwhile, MinHashLSH-based techniques readily admit parameter tuning with relatively low runtime and memory costs.
 Furthermore, MinHashLSH and LSHBloom both benefit from established theoretical analysis that effectively bounds their false positive and false negative rates given the values for the Jaccard similarity threshold and number of permutations \cite{leskovec2014finding}.
 MinHashLSH-based techniques offer the greatest opportunity to tune parameters for the best deduplication performance, particularly when faced with a constrained parameter tuning budget.
 This is particularly true for LSHBloom, which is significantly cheaper to run than MinHashLSH in terms of wall clock time and disk usage---particularly as we increase the size of our dataset to several million or several billion documents (see \autoref{sec:scaling}).

\begin{figure*}[ht!]
    \includegraphics{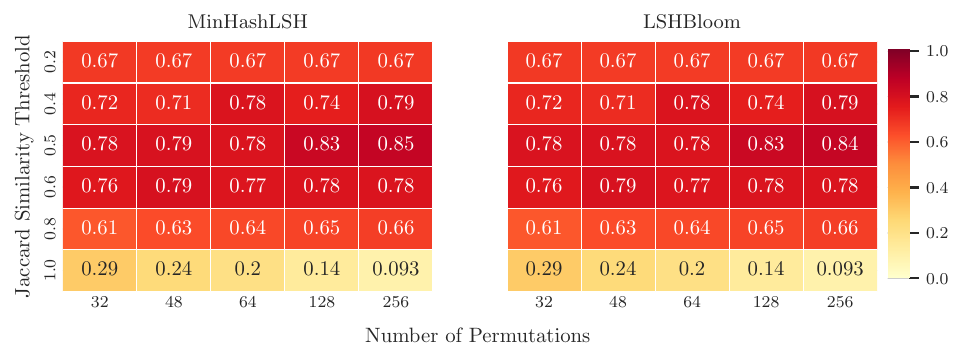}
    \caption{F1 score for LSH techniques as a function of number of permutations (x-axis) and Jaccard similarity threshold (y axis).}
    \label{fig:lsh_heatmap}
\end{figure*}

\subsection{Hyperparameter Tuning}



\subsubsection{LSH Methods}
The parametrization of methods such as MinHashLSH and LSHBloom has dramatic effects on their accuracy, so tuning the parameters is a critical aspect of deploying them for large-scale deduplication efforts. \autoref{fig:lsh_heatmap} shows the F1 score for MinHashLSH and LSHBloom, respectively, with different settings for the Jaccard similarity threshold, $T$, and the number of permutation functions for MinHashing on the tuning dataset. As the number of permutation functions increases, the MinHash estimate of the Jaccard similarity becomes more accurate, and performance generally improves, which is consistent with the theory of MinHashing~\cite{broder1998min}. Correspondingly, we use 256 MinHash permutations in our subsequent experiments. 
The threshold parameter affects the F1 score the most, as it defines the sensitivity of the deduplication algorithm to overlap in text content. For our particular benchmark, a threshold of 0.5 yields the best performance; more stringent thresholds lead to a diminishing F1 score.

\begin{figure*}[ht!]
     \includegraphics{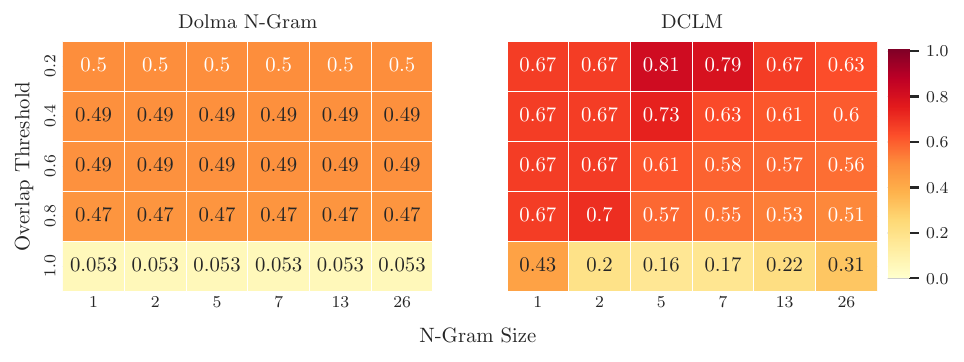}
    \caption{F1 Score for N-Gram techniques as a function of N-Gram size (x axis) and overlap threshold (y axis).}
    \label{fig:ngram_heatmaps}
\end{figure*}

\subsubsection{Paragraph Level and N-Gram Methods}
 
\autoref{fig:ngram_heatmaps} shows the results of varying both the n-gram size and overlap threshold parameters for each n-gram method. DCLM is fairly successful, even approaching the F1 score of the MinHashLSH-based techniques, and demonstrates moderate sensitivity to the overlap threshold parameter. Smaller n-gram sizes seem to perform better as well---likely due to the increased granularity with which they enable comparison between a document and the remainder of the corpus. For DCLM, the best setting is using an overlap threshold of 0.2 and an n-gram size of 5. In contrast, Dolma-Ngram seems to be fairly insensitive to the choice of parameters and performs relatively poorly. The difference between these two very similar techniques may stem from DCLM's use of the UniSeg tokenizer, which splits text into meaningful units based on rules defined by the Unicode standard---Dolma-Ngram simply splits text by whitespace. 

\begin{figure}[ht!]
    \centering
    \includegraphics{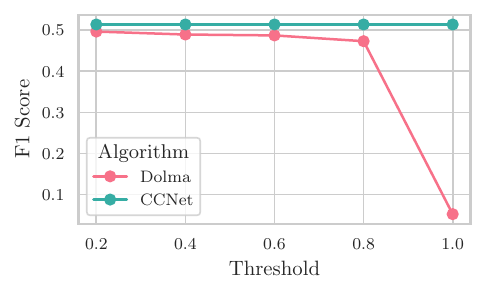}
    \caption{F1 score vs.\ threshold, paragraph-level techniques.}
    \label{fig:correctness_para_f1_vs_threshold}
\end{figure}

Figure \ref{fig:correctness_para_f1_vs_threshold} shows the impact of adjusting the overlap threshold parameter for paragraph-level methods, DOLMA and CCNet, on F1 score.
Predictably, distinguishing documents at the granularity of individual paragraphs is highly error-prone and results in poor performance. The best F1 for paragraph-level methods is achieved by setting a low threshold of 0.2.
While n-gram methods enable finer-grained comparison between individual documents and the rest of the corpus, these methods evaluate overlap by determining the portion of n-grams in a document that is exactly duplicated in the corpus. 
This is somewhat similar to the notion of evaluating the set overlap between documents, but can be sensitive to the relative frequency of n-grams within a document, resulting in increased error.
 Moreover, paragraph-level techniques demonstrate a lack of sensitivity to changes in their parameter settings in this benchmark, indicating bias in their predictions.
 Specifically, we find that paragraph-level methods tend to be biased toward marking documents as duplicates.

 Table~\ref{tab:tuning_results} shows the parameter settings for each technique that we use for the remainder of our experiments.

\begin{table}[ht!]
\caption{Best settings for each deduplication technique}


\label{tab:tuning_results}
\centering
\begin{tabular}{ c | c | c }
 \textbf{Technique} & \textbf{N-gram Size}  & \textbf{Threshold} \\ 
 \hline
 MinHashLSH & 1 & 0.5 \\  
 LSHBloom & 1 & 0.5 \\
 Dolma-Ngram & 5 & 0.2 \\
 DCLM & 5 & 0.2 \\
 Dolma & - & 0.2 \\
 CCNet & - & 0.2 \\
\end{tabular}
\end{table}

\subsection{Deduplication Fidelity}
\label{sec:50k}

\begin{figure*}[ht!]
    
    \includegraphics{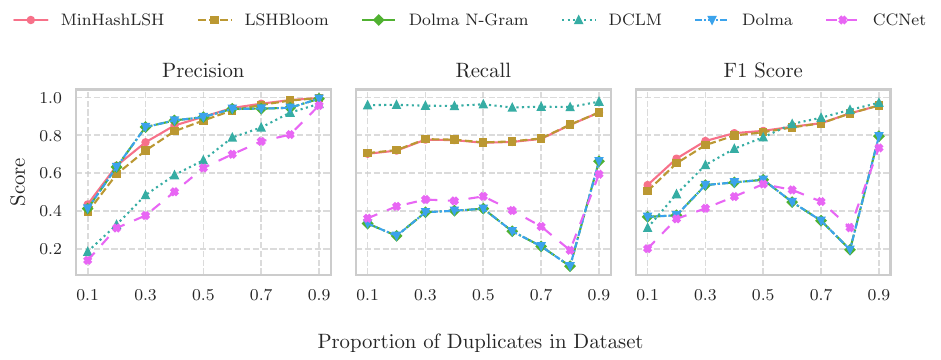}
    \caption{Fidelity metrics for deduplication techniques for various levels of duplication.}
    \label{fig:imbalance_comparison}
\end{figure*}
 
\subsubsection{Deduplication Quality}
Here we evaluate the fidelity of each technique using metrics such as precision, recall, and F1 score. We do so by constructing datasets of 50,000 documents each with varying levels of duplication between 10\% and 90\%. We construct these datasets using the procedure described in \autoref{sec:25k}. 

\autoref{fig:imbalance_comparison} illustrates the precision, recall, and F1 score of each technique as we vary the proportion of duplicate documents in our dataset. Notably, in terms of F1 score, both MinHashLSH and LSHBloom outperform all other methods---only slightly outperformed by DCLM and Dolma-Ngram (whose performance tracks closely together) when the vast majority ($>$60\%) of the dataset consists of duplicates, highly unlikely in real-world scenarios. MinHashLSH and LSHBloom also achieve nearly identical F1 scores across experimental settings, exemplifying that our technique can closely match MinHashLSH's performance while affording considerable gains in scalability (to be shown below in \autoref{sec:scaling_experiments}).

Looking more closely, MinHashLSH and LSHBloom tend to consistently outperform the other techniques in terms of precision, which is particularly noteworthy for those settings where duplicates are rare, such as when duplicates only comprise 10 or 20\% of the dataset. The next best technique in terms of precision is Dolma's paragraph-level deduplication technique, with its precision tracking closely with that of the MinHash-based techniques. However, Dolma's dismal recall scores ($<0.4$) suggest that this is likely the result of under-identifying duplicate instances rather than being a reflection of greater capacity for discernment. DCLM, Dolma-Ngram, and CCNet tend to underperform the other techniques in terms of precision. Note that there is a natural tendency for precision scores to increase as the proportion of duplicates in the dataset increases. This is due to a greater number of true positives being present in the dataset. 

In terms of recall, DCLM and Dolma-Ngram tend to perform the best, although their relatively stable recall scores come at the cost of decreased precision, particularly when duplicates are rare. This indicates that, while these techniques identify most duplicates, they struggle to differentiate between duplicate and non-duplicate entities. Meanwhile, MinHashLSH and LSHBloom provide virtually identical recall scores to one another between 0.7 and 0.9 while simultaneously demonstrating strong precision. Paragraph-level techniques demonstrate lower recall than all other techniques by a wide margin. 

Our experimental results indicate that LSHBloom achieves deduplication performance comparable to the state-of-the-art---indeed, LSHBloom achieves essentially identical performance to MinHashLSH, one of the most popular choices for text deduplication, with only a marginal degradation in precision.

\begin{figure*}[ht!]
    \begin{subfigure}{0.5\textwidth}
        \includegraphics{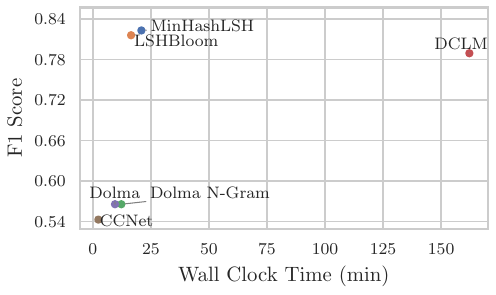}
        \caption{Quality vs.\ Wall Clock Time}
        \label{fig:pareto_runtime}
    \end{subfigure}
    \hspace{-1.5mm}
    \begin{subfigure}{0.5\textwidth}
        \includegraphics{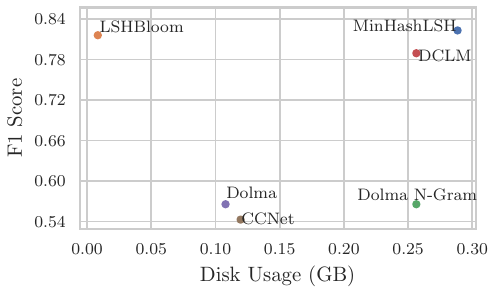}
            \caption{Quality vs.\ Disk}
            \label{fig:pareto_disk}
    \end{subfigure}
    \caption{Pareto plots for F1 score vs.\ resource usage.} 
    \label{fig:pareto}
\end{figure*}

\subsubsection{Performance vs.\ Quality Tradeoffs}
Here we investigate tradeoffs between the costs of various deduplication techniques (in terms of runtime and disk usage) and their deduplication performance. We visualize data from our evaluation of the balanced dataset of 50,000 documents. 

Figures~\ref{fig:pareto_runtime} and~\ref{fig:pareto_disk} plot the F1 score for each deduplication technique against runtime and disk usage, respectively. MinHashLSH and LSHBloom dominate the alternative techniques, achieving the highest F1 scores while also exhibiting competitive runtime. LSHBloom runs faster than MinHashLSH and this runtime advantage becomes more exaggerated at scale (see \autoref{sec:scaling_experiments}). LSHBloom also displays a comparable F1 score to traditional MinHashLSH while using just a fraction of the disk space for its index. As shown below in \autoref{sec:scaling_experiments}, this is a crucial distinction, as rapidly increasing disk usage renders deduplication efforts infeasible at the scale of many millions or billions of documents.

Our results indicate that LSHBloom intelligently discerns between duplicate and non-duplicate documents (as demonstrated by its high precision scores) and can adeptly identify the majority of duplicate documents (as demonstrated by its strong recall scores). Our results also illustrate that the main difference in performance between MinHashLSH and LSHBloom comes in the form of slightly reduced precision for the latter. Intuitively, this makes sense as LSHBloom yields slightly more false positive predictions due to its use of the Bloom filter. Overall, however, both yield similar F1 scores. This, combined with LSHBloom's greatly reduced resource usage compared to MinHashLSH, positions it as the best choice for text deduplication.

\subsection{Deduplication Resource Usage at Scale}
\label{sec:scaling_experiments}

As LLM training datasets grow ever larger, it is important that text deduplication strategies can scale effectively to larger 
quantities of documents. 
Here we work with the peS2o dataset \cite{peS2o} of roughly 39M documents.
We evaluate each deduplication method on subsets of peS2o of various sizes and compare them in terms of wall clock time and disk usage (e.g., for storing an index or Bloom filter). We do not evaluate the n-gram techniques for scaling capacity on peS2o due to their prohibitive cost. DataComp-LM, for example, requires almost three hours just to deduplicate 50,000 examples in our previous benchmark in \autoref{sec:50k}. Even processing just 10\% of peS2o using this technique would require over 200 hours by extrapolation. We find that Dolma-Ngram is similarly prohibitive in runtime, likely due to the time cost of processing many n-grams in documents that are frequently much larger than those in our previous benchmark. We discuss the results for other methods below.

\begin{figure*}[ht!]
    \begin{subfigure}{0.5\textwidth}
        \includegraphics{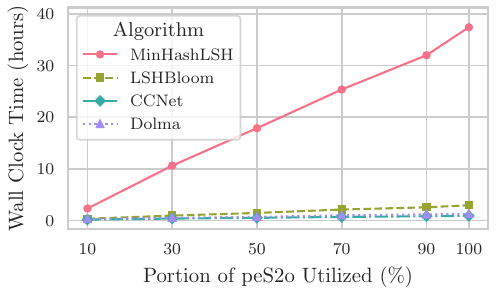}
        \caption{Wall clock times over subsets of peS2o.}
        \label{fig:wall_clock_time}
    \end{subfigure}%
    \begin{subfigure}{0.5\textwidth}
        \includegraphics{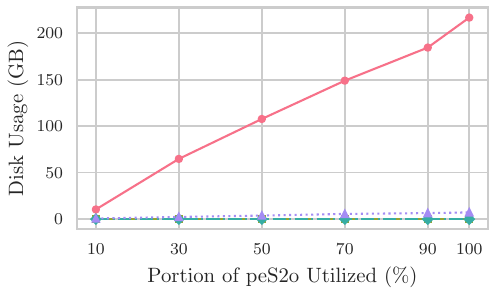}
        \caption{Disk Usage over subsets of peS2o.}
        \label{fig:disk_usage}
    \end{subfigure}
    \caption{Resource usage at scale for each deduplication technique.}
    \label{fig:scaling_comparison}
\end{figure*}

\begin{figure}[ht!]
    \centering
    \includegraphics{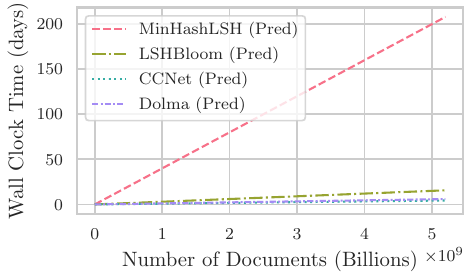}
    \caption{Extrapolated wall clock times per method at scale.}
    \label{fig:time_extrapolation}
\end{figure}


\subsubsection{Scaling of Resource Usage}
\label{sec:scaling}

In this section, we study how resource usage grows with the number of documents. In these experiments, we use the same settings shown in \autoref{tab:tuning_results}. We set the effective false positive rate for all Bloom filters to 1e-10.

\autoref{fig:wall_clock_time} shows the wall clock time to deduplicate various subsets of peS2o with each method.
Notably, MinHashLSH incurs the largest runtime while paragraph-level techniques such as DOLMA and CCNet are relatively fast. Meanwhile, LSHBloom, which has comparable deduplication performance to MinHashLSH, is significantly faster---demonstrating comparable runtime to that of a simpler paragraph-level technique. This is due to our Bloom filter index, for which insertion and querying only require a modest number of integer hashing operations. This speedup is illustrated in \autoref{fig:wall_clock_time_breakdown}. To deduplicate all of peS2o, LSHBloom takes just three hours, whereas MinHashLSH takes over 37 hours---over a 12$\times$ speedup.


\autoref{fig:disk_usage} shows the disk usage for each method. We again see a similar trend.  The paragraph-level techniques maintain only a single Bloom filter, which leads to extremely low disk usage. LSHBloom's index also grows linearly in the number of documents, but at a reduced rate compared to that of MinHashLSH. When processing all of peS2o, LSHBloom requires just 11~GB of disk while MinHashLSH uses over 200~GB of disk---an 18$\times$ reduction.

\subsubsection{Estimated Resource Usage at Scale}


Our results show that LSHBloom has clear benefits in efficiency over MinHashLSH, even as we increase the number of documents we process.
Because runtime scales approximately linearly for each method, we model runtime as a linear function of the number of documents. Using this model, we extrapolate to estimate each method’s resource requirements when processing approximately 5B documents.
\autoref{fig:time_extrapolation} shows the extrapolated runtime for each technique against corpus size in the number of documents.
MinHashLSH would require 200 days to process 5B documents.
Meanwhile, LSHBloom could achieve the same in only 15 days: a speedup of over 13$\times$. These figures do not account for parallelization, which could be applied to both MinHashLSH and LSHBloom---for instance, by splitting the dataset into subsets for processing and progressively aggregating each reduced subset.

We also consider disk usage, a crucial limiting factor. If index sizes grow beyond what is feasible to store, read, and write, then deduplication efforts would be infeasible at internet-scale.
\autoref{tab:lsh_extrapolated_disk_comparison} shows the linearly extrapolated index size for MinHashLSH and the computed index size for LSHBloom at various effective false positive rate settings, $p_\textsubscript{effective}$, for a corpus of both 5B and 100B documents. We compute the size of the LSHBloom index exactly using the method described in \autoref{sec:lshbloom_advantage}.
For 5B documents, MinHashLSH requires just over 277~TB.
By contrast, even an extremely conservative effective false positive rate of 1/5B for LSHBloom yields an index of only 15.5~TB.
At this scale, LSHBloom offers an almost 18$\times$ space savings over the traditional MinHashLSH algorithm. Even for an extreme-scale corpus of 100B documents (on the order of a dataset such as RedPajama), MinHashLSH would require 555~TB of disk space, whereas LSHBloom with a conservative $p\textsubscript{effective}$ = 1/100B would only require 32~TB---a 17$\times$ space advantage with near-zero false positive error overhead. Of course, the user may choose to increase the false positive rate in exchange for greater space savings.

\begin{center} 
\begin{minipage}{\columnwidth}
\centering
\small 
\captionsetup{type=table} 
\caption{Comparison of Extrapolated Index Storage Requirements} 
\label{tab:lsh_extrapolated_disk_comparison}
\vspace{-5pt} 

\begin{tabular}{@{}llcc@{}}
\toprule
\textbf{Technique} & \textbf{Bloom FP} & \multicolumn{2}{c}{\textbf{Index Disk Usage (TB)}} \\ \cmidrule(l){3-4} 
                   & \textbf{Overhead} & \textbf{$N=5 \times 10^9$} & \textbf{$N=10^{11}$} \\ \midrule
MinHashLSH         & ---               & 277.68               & 555.35               \\ \addlinespace
\textbf{LSHBloom}  & $10^{-5}$         & 8.33                 & 16.66                \\
                   & $10^{-8}$         & 12.11                & 24.21                \\
                   & $1/N$             & 15.50                & 31.76                \\ \bottomrule
\end{tabular}
\end{minipage}
\end{center}

\vspace{-3pt}

\section{Conclusions}

We have introduced LSHBloom, a high-performance technique for internet-scale text deduplication.  
We contribute one of the first parameter tuning studies for several state-of-the-art text deduplication algorithms. 
We also conducted a thorough comparison of deduplication fidelity between state-of-the-art deduplication algorithms commonly used in LLM data pipelines, using their best settings on a custom synthetic benchmark of 50,000 scientific documents. We found that even following extensive parameter tuning of other methods, LSHBloom achieves state-of-the-art deduplication fidelity, with an F1 score within 1\% of that achieved by MinHashLSH, uses far less storage, and achieves significantly higher throughput. On the peS2o dataset of 39M documents, we showed that LSHBloom can feasibly deduplicate millions of records due to its small disk footprint (18$\times$ less than MinHashLSH) and high throughput (12$\times$ faster than MinHashLSH). Alternatives demonstrate either poor deduplication fidelity, inferior throughput, or prohibitive storage requirements. 
In summary, LSHBloom provides both the powerful deduplication fidelity of MinHashLSH and the scaling capacity of cheaper, heuristic deduplication algorithms.

In the future, we will further enhance LSHBloom's throughput by carefully employing parallelization over subsets of text datasets when inserting them into our index. Additionally, though very useful, our synthetic benchmark has drawbacks---namely, in that it mainly emulates forms of duplication native to web scraping-based ingestion. Ideally, future work will involve benchmark datasets that cover more diverse forms of data duplication. 



\bibliographystyle{ACM-Reference-Format}
\bibliography{refs}

\end{document}